\documentclass[preprint,5p,times,twocolumn]{elsarticle}
\usepackage{natbib}
\usepackage{multirow}
\usepackage{color}
\usepackage{multicol}
\usepackage{graphicx}
\usepackage[table]{xcolor}

\usepackage{amssymb}
\usepackage{gensymb}
\usepackage{fontenc}
\usepackage{amsmath}
\usepackage{hyperref}

\usepackage{lineno}

\newcommand{\argmaxA}{\mathop{\mathrm{arg\,max}}}

\begin{document}

\begin{frontmatter}

\title{3D Inception-based CNN with sMRI and MD-DTI data fusion for Alzheimer's Disease diagnostics}
	
\author[msu]{Alexander Khvostikov\corref{cor1}}
\ead{xubiker@gmail.com}

\author[labri,maroc]{Karim Aderghal}
\ead{aderghal.karim@gmail.com}

\author[msu]{Andrey Krylov}
\ead{kryl@cs.msu.ru}

\author[bordeaux2]{Gwenaelle Catheline}
\ead{gwenaelle.catheline@u-bordeaux.fr}

\author[labri]{Jenny Benois-Pineau}
\ead{jenny.benois-pineau@u-bordeaux.fr}

\author{for the Alzheimer's Disease Neuroimaging Initiative\tnoteref{adni}}

\cortext[cor1]{Corresponding author}

\tnotetext[adni]{Data used in preparation of this article were obtained from the Alzheimer's Disease Neuroimaging Initiative (ADNI) database (\url{http://adni.loni.usc.edu}). As such, the investigators within the ADNI contributed to the design and implementation of ADNI and/or provided data but did not participate in analysis or writing of this report. A complete listing of ADNI investigators can be found at: \url{http://adni.loni.usc.edu/wp-content/uploads/how_to_apply/ADNI_Acknowledgement_List.pdf}.}

\address[msu]{Lomonosov Moscow State University, Department of Computational Mathematics and Cybernetics, Moscow, Russia}

\address[labri]{LaBRI UMR 5800, University of Bordeaux, Bordeaux, France}

\address[maroc]{LabSIV, University Ibn Zhor, Agadir, Morocco}

\address[bordeaux2]{EPHE, PSL, UMR5287, Institut de Neurosciences Cognitives et Int\'egratives d'Aquitaine, Bordeaux, France}

\begin{abstract}
In the last decade, computer-aided early diagnostics of Alzheimer’s Disease (AD) and its prodromal form, Mild Cognitive Impairment (MCI), has been the subject of extensive research. Some recent studies have shown promising results in the AD and MCI determination using structural and functional Magnetic Resonance Imaging (sMRI, fMRI), Positron Emission Tomography (PET) and Diffusion Tensor Imaging (DTI) modalities. Furthermore, fusion of imaging modalities in a supervised machine learning framework has shown promising direction of research. 

In this paper we first review major trends in automatic classification methods such as feature extraction based methods as well as deep learning approaches in medical image analysis applied to the field of Alzheimer's Disease diagnostics. Then we propose our own design of a 3D Inception-based Convolutional Neural Network (CNN) for Alzheimer's Disease diagnostics. The network is designed with an emphasis on the interior resource utilization and uses sMRI and DTI modalities fusion on hippocampal ROI. The comparison with the conventional AlexNet-based network using data from  the  Alzheimer’s  Disease Neuroimaging  Initiative  (ADNI)  dataset (\url{http://adni.loni.usc.edu}) demonstrates significantly better performance of the proposed 3D Inception-based CNN.
\end{abstract}

\begin{keyword}
Medical Imaging, Alzheimer’s Disease, Mild Cognitive Impairment, Machine Learning, Deep learning, Convolutional Neural Networks, Data Fusion.
\end{keyword}

\end{frontmatter}


\section{Introduction}
\label{S:Introduction}

Alzheimer’s Disease (AD) is the most common type of dementia. It is characterized by degeneration of brain cells which results in changes of brain structures noticeable on images form different imaging modalities e.g. sMRI, DTI, PET. Brain development and aging are key topics in neuroscience.
The study of normal brain maturation and age-related brain atrophy is crucial to better understand normal brain development and a large variety of neurological disorders \cite{coupe2017towards}, which also include AD. With the development of machine learning approaches, research on computer-aided diagnostics (CAD) has become very much intensive \cite{volumetric:VBM},\cite{c2.2_3D_CNN_USA, c4_sparse_encoder_Korea, c8_medical_UK, c5_demnet_Philippines}. 

Images of different modalities such as structural and functional magnetic resonance imaging (sMRI, fMRI), positron emission tomography (PET) and diffusion tensor imaging (DTI) scans  can be used for early and non-invasive detection of Alzheimer's Disease.

The majority of earlier works were focused on the volumetric approaches that perform comparison of anatomical brain structures assuming one-to-one correspondence between subjects. The wide-spread voxel-based morphometry (VBM) \cite{volumetric:VBM} is an automatic volumetric method for studying the differences in local concentrations of white and gray matter and comparison of brain structures of the subjects to test with reference normal control (NC) brains. Tensor-based morphometry (TBM) \cite{volumetric:TBM} was proposed to identify local structural changes from the gradients of deformation fields when matching tested brain and the reference healthy NC. Object-based morphometry (OBM) \cite{volumetric:OBM} was introduced for shape analysis of anatomical structures.

In general, the automatic classification on brain images of different modalities can be applied to the whole brain \cite{c2.2_3D_CNN_USA, c4_sparse_encoder_Korea, c8_medical_UK, c5_demnet_Philippines},
or performed  using the domain knowledge on specific regions of interest (ROIs). Structural changes in some  brain structures e.g. hippocampal ROI are strongly correlated to the disease \cite{pierrick}. The changes in such regions are considered as AD biomarkers.  

Advances in computer vision and content-based image retrieval research made penetrate the so-called feature-based methods into classification approaches for AD detection \cite{f1_Jenny_MKL,f3_Jenny_pcc,f5_Jenny}. 
The reason for this is in inter-subject variability, which is difficult to handle in VBM. On the contrary,  the quantity of local features, which can be extracted from the brain scans, together with captured particularities of the image signal allowed a more efficient classification with lower computational workload \cite{f5_Jenny,hippo_texture} compared to VBM and OBM methods.

Lately with the development of neural networks the feature-based approach became less popular and is being gradually replaced with convolutional neural networks of different architectures \cite{survey2017}.

In the present paper we continue our previous work \cite{self_arxiv}. We give a substantial overview of recent trends in classification of different brain imaging modalities in the problem of computer-aided diagnostics of Alzheimer's Disease and its prodromal stage, i.e. mild cognition impairment (MCI) and propose a new algorithm for this purpose. The algorithm is based on the recent trend in supervised machine learning such as Deep Convolutional Neural Networks (CNNs), and its specific architecture known as "Inception" \cite{inception}.

Our contribution is in the design of a new 3D Inception-based convolutional neural network architecture, based on the idea of improved utilization of the computer resources inside the network by reducing the number of network parameters without sacrificing the classification performance. We compare the proposed network with the conventional AlexNet-based network \cite{alexnet} and demonstrate better performance of the proposed CNN. The network uses 3D volumes of hippocampal ROIs as input. Furthermore, two modalities are fused: sMRI and DTI allowing for joint exploration of different modalities where structural changes of hippocampal ROI are observed by medical experts. In our work we use a subset of ADNI database (\url{http://adni.loni.usc.edu}) with two image modalities (sMRI and DTI) available for the same cohort of patients.

The paper is organized as follows. In Section \ref{sec:Classification Approaches} we overview the recent trends in classification of brain images in the problem of AD detection. Main feature-based approaches are presented in Section  \ref{sec:texture_based}. In section \ref{sec:networks} we compare different approaches based on neural networks. Particular attention is paid in each case to fusion of modalities. All reviewed approaches are compared in Table \ref{table:comparison}. In section \ref{sec:network} we present the new 3D CNN architecture. In section \ref{sec:Experiments_and_results} the implementation details of the proposed algorithm are described and the main results are presented. Sections \ref{sec:Discussion}, \ref{sec:Conclusion} contain discussion and conclusion of our work and outline research perspectives. 

\section{Review of the existing classification methods in the problem of AD detection}
\label{sec:Classification Approaches}
As an alternative to heavy volumetric methods, feature-based approaches were applied in the problem of AD detection using domain knowledge both on the ROI biomarkers and on the nature of the signal in sMRI and DTI modalities which is blurry and cannot be sufficiently well described by conventional differential descriptors such as SIFT\cite{SIFT} and SURF\cite{SURF}. 

\subsection{Feature-based classification}
\label{sec:texture_based}

Feature-based classification can be performed on images of different modalities. Here we compare and discuss the usage of sMRI, DTI and sMRI fusion with other modalities.

\subsubsection{sMRI}

In previous joint work \cite{f3_Jenny_pcc}, \citeauthor{f3_Jenny_pcc} computed local features on  sMRI scans in hippocampus and posterior cingulate cortex (PCC) structures of the brain. The originality of the work consisted in the usage of Gauss-Laguerre Harmonic Functions (GL-CHFs) instead of traditional SIFT\cite{SIFT} and SURF\cite{SURF} descriptors. CHFs perform image decomposition on the orthonormal functional basis, which allows capturing local directions of the image signal and intermediate frequencies. It is  similar to Fourier decomposition, but is more appropriate in the case of smooth contrasts of sMRI modality. For each projection of each ROI a signature vector was calculated using a bag-of-visual-words model (BoVWM) with a low-dimensional dictionary with  300 clusters. This led to the total signature length of 1800 per image. Principal component analysis was then applied to reduce the signature length to 278. The signatures then were classified using SVM with RBF kernel and 10-fold cross-validation and reached the accuracy level of 0.838, 0.695, 0.621 for AD/NC, NC/MCI and AD/MCI binary classification problems accordingly on the subset of ADNI database.

In \cite{salvatore2015magnetic} the authors used a feature-based approach for AD diagnostics with an emphasis on the distinguishing the convertible and non-convertible MCI stages. Although, the performance of the proposed method was not very high (0.76 for AD/NC classification of the subjects from ADNI database), the authors demonstrated that hippocampus, entorhinal cortex, basal ganglia, gyrus rectus, precuneus, and cerebellum regions have a strong influence on the classification of the pre-clinical AD stages.

The search of the brain areas correlated with AD is also the base of the method proposed by \citeauthor{landmark_AD} in \cite{landmark_AD}. The authors used morphological features to identify brain regions with significant difference for AD and NC subjects. This morphological features were further classified with an SVM classifier. The authors have achieved the 0.837 accuracy level for AD/NC classification performed on the subset of ADNI database.

Paper \cite{AD_patch} also deserves a special attention as the authors presented a patch-based descriptor that encodes local displacements due to atrophy between a pair of longitudinal MRI scans. The conventional logistic regression classifier with the proposed descriptor achieved 0.76 accuracy in predicting the MCI converters (MCI patients that lately converted to AD). Two hundred and sixty four subjects including both non-converter and converter MCI samples from ADNI dataset were selected to evaluate the proposed method.

\subsubsection{DTI}

This modality is probably the most recent to be used for AD classification tasks. Both  Mean Diffusivity (MD) and Fractional Anisotropy (FA) maps are being explored for this purpose.  
In \cite{g1_graph_ensemble_2017} the authors acquired DTI images of 15 AD patients, 15 MCI patients, and 15 healthy volunteers (NC). After the preprocessing steps the FA map, which is an  indicator of brain connectivity,  was calculated. The authors considered 41 Brodmann areas, calculated the connectivity matrices for this areas and generated a connectivity graph with corresponding 41 nodes. Two nodes corresponding to Brodmann areas are marked with an edge if there is at least one fiber connecting them. Then the graph is described with the vector of features, calculated for each node and characterizing the connectivity of the node neighborhood. Totally each patient is characterized by 451 features. The vectors were reduced to the size of 430 and 110 using ANOVA-based feature selection approach. All vectors were classified with the ensemble of classifiers (Logistic regression, Random Forest, Gaussian native Bayes, 1-nearest neighbor, SVM) using 5-fold cross-validation. The authors have achieved the 0.8, 0.833, 0.7 accuracy levels for AD/NC, AD/MCI and MCI/NC accordingly on their custom database.

Another methodology is described in \cite{g3_svm_DTI_Korea}. The authors use the fractional anisotropy (FA) and mode of anisotropy (MO) values of DTI scans of 50 patients from the LONI Image Data Archive (\url{https://ida.loni.usc.edu}). After non-linear registration to the standard FA map, the authors calculate the skeleton of the mean FA image as well as MO and perform the second step of registration. After that a Relief feature algorithm is performed on all voxels of the image, relevant ones are used for 10-cross validation training the SVM classifier with RBF kernel. The declared accuracy is 0.986 and 0.977 for classification  AD/MCI, AD/NC accordingly.

\subsubsection{Data fusion}

In \cite{g2_sift_China} the authors use a fusion of sMRI and PET images together with canonical correlation analysis (CCA). After preprocessing and aligning images of 2 modalities given the covariance data of sMRI and PET images they find the projection matrices by maximizing the correlation between projected features. Here
\[
X_1 \in R^{d \times n}, X_2 \in R^{d \times n}
\]
are the \(d\)-dimentional sMRI and PET features of \(n\) samples,
\[
\Sigma = \begin{bmatrix}
    \Sigma_{11} &  \Sigma_{12} \\
    \Sigma_{21} &  \Sigma_{22} \\
  \end{bmatrix}
\]
is a covariance matrix, 
\[
(B_1, B_2) = \argmaxA_{(B_1, B_2)} \frac{B_1^T \Sigma_{12} B_2}{\sqrt[]{B_1^T \Sigma_{11} B_1}\sqrt[]{B_2^T \Sigma_{22} B_2}}
\]
are the projection matrices and
\[
Z_1 = B_1^T X_1, Z_2 = B_2^T X_2
\]
are the resulting projections.
The authors construct the united data representation for each patient:
\[
F = [X_1; X_2; Z_1; Z_2] \in R^{4d \times n}
\]
and calculate SIFT descriptors. This descriptors are used to form the BoVW model, the classification is performed using SVM. The achieved accuracy is 0.969, 0.866 and for classifying AD/NC and MCI/NC accordingly on a subset of ADNI database.

\citeauthor{f5_Jenny} in \cite{f5_Jenny} demonstrated the efficiency of using the amount of cerebrospinal fluid (CSF) in the hippocampal area calculated by an adaptive Otsu's  thresholding method as an additional feature for AD diagnostics. In \cite{f1_Jenny_MKL} they further improved the result of \cite{f3_Jenny_pcc} by combining visual features derived from sMRI and DTI MD maps with a multiple kernel learning scheme (MKL). Similar to \cite{f3_Jenny_pcc} they selected hippocampus ROIs on the axial, saggital and coronal projections and described them using Gauss-Laguerre Harmonic Functions (GL-CHFs). These features are clustered into 250 and 150 clusters for sMRI and MD DTI modalities and encoded using the BoVW model. Thus they got three sets of features: BoVW histogram for sMRI, BoVW histogram for MD DTI and CSF features. The obtained vectors are classified using MKL approach based on SVM. The achieved accuracy is 0.902, 0.794, 0.766 for AD/NC, MCI/NC and AD/MCI classification on a subset of ADNI database.

\subsection{Classification with neural networks}
\label{sec:networks}

Deep neural networks (DNNs) and specifically convolutional NN (CNNs) have become popular now due to their good generalization capacity and available GPU Hardware needed for parameter optimization. Despite the fact that CNNs originally were applied for general purpose images, they are becoming a wide-spread methodology in medical image analysis as well \cite{survey2017}.It should be also noted, that the application area of CNNs is not limited with the direct AD prediction. For example, in \cite{landmark_CNN} the authors used a combination of two CNNs to perform anatomical landmarks detection. In \cite{landmark_path_2017} the authors designed a CNN-based method which is able not only to detect anatomical landmarks but also find the optimal path to the target object in the volumetric space. \citeauthor{flair} in \cite{flair} used a 3D Fully Convolutional network to predict missing pulses in a fluid-attenuated inversion recovery (FLAIR) MRI pulse sequence.

Their main drawback for AD classification is the small amount of available training data and also a low resolution of input images when the ROIs are considered. Although, some studies try to offer a comprehensive analysis of the available data and describe it with optimizes models \cite{coupe2017towards}, still the amount of available data is one of the main problems in case of AD diagnostics.  This problem in the context of CNNs can be eliminated in several ways: i) by using shallow networks with relatively small number of neurons, ii) applying transfer learning from an  existing trained network or iii) pretraining  some of the layers of the network.

Forming shallow networks kills the idea of deep learning to recognize structures at different scales and reduces the generalization ability of the network, so this methodology has not often been used since recently, despite it has shown decent results \cite{aderghal2017classification}. In this case the classification performance could be enhanced by selecting several ROIs in each image and applying the voting rule. In particular in \cite{ex_2} authors used 7 ROIs in each sMRI image.

One way to enlarge the dataset is to use domain-dependent data augmentation. In the case of medical images this often comes down to mirror flipping, small-magnitude translations and weak Gaussian blurring \cite{aderghal2017classification}.

Another way is to use more input data e.g. consider several ROIs instead of one. So \citeauthor{ex_6} in \cite{ex_6} first identify discriminative 50 anatomical landmarks from MR images in a data-driven manner, and then extract multiple image patches around these detected landmarks. After that they use a deep multi-task multi-channel convolutional neural network for disease classification. The authors addressed the problem of classification of patients into NC, stable MCI (sMCI) an progressive MCI (pMCI). The authors used MRI images from ADNI database containing in total 1396 images and achieved 0.518 accuracy in four-class (NC/sMCI/pMCI/AD) classification.

A more simplified idea was proposed by \citeauthor{ex_7} in \cite{ex_7} as they used a number of 3D convolutional neural networks with 4 layers together with late fusion. With the subset of ADNI database of 428 sMRI images the authors achieved an accuracy value of 0.872 for AD/NC classification.

A nearly similar approach was proposed in \cite{liu2018multi}. The authors designed a set of local multimodal (sMRI + PET) 3D CNNs, each of them processes a small patch of the whole brain image. Then a set of upper high-level CNNs are cascaded to combine the features learned from local CNNs and learn the latent multimodal features for image classification.

One more problem that is specific for multimodal solutions is the incompleteness of the multimodal data as not all data can be collected for every individual. The multi-task approach proposed in \cite{multi_incomplete} was designed to solve this problem. \citeauthor{multi_incomplete} used a CNN with two inputs, corresponding to the images of first and second modalities, and three outputs, corresponding to the three classification tasks (subjects with first modality, subjects with second modality or subjects with both modalities). The subnet for each task was trained separately. The authors declared the accuracy level of 0.658 for ternary AD/MCI/NC classification using sMRI and PET images from a subset of ADNI database.

\subsubsection{Autoencoders}

\begin{figure}[h]
\centering\includegraphics[width=0.7\linewidth]{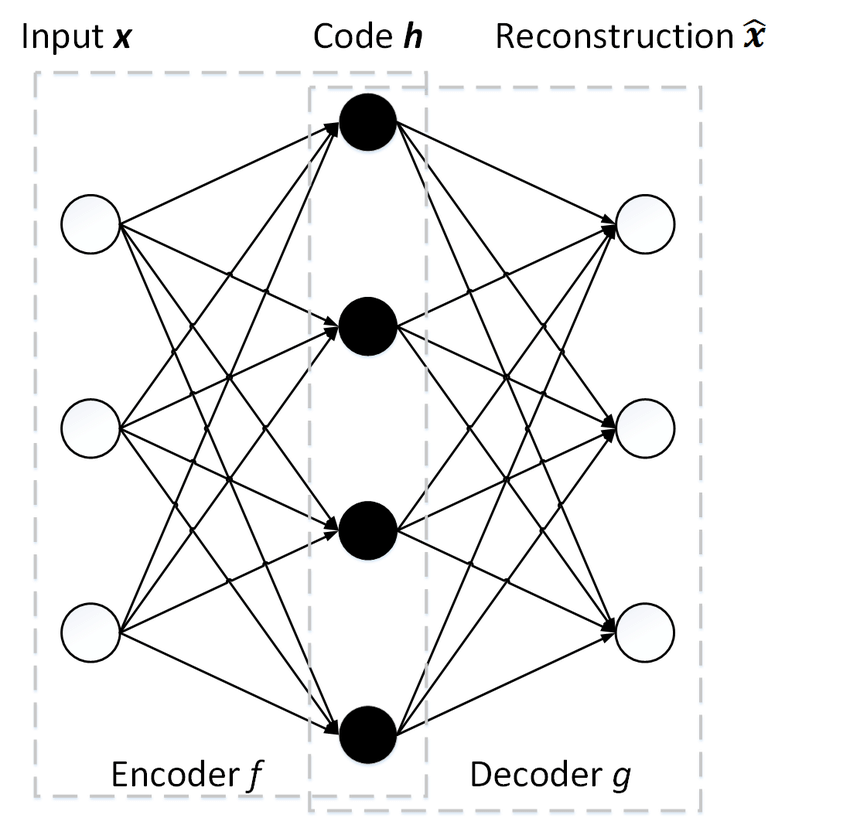}
\caption{Architecture of an autoencoder}
\label{fig:autoencoder}
\end{figure}

The idea of pretraining some of the layers in the network is easily implemented with autoencoders (AE) or in  image processing tasks more often with convolutional autoencoders (CAE). Autoencoder consists of an input layer, hidden layer and an output layer, where the input and output layers have the same number of units (Fig.\ref{fig:autoencoder}). Given  the input vector \(x \in \mathbb{R}^n \) autoencoder maps it to the hidden representation \(h\):
\[
h = f(Wx+b),
\]
where \(W \in \mathbb{R}^{p \times n} \) are the weights, \(b \in \mathbb{R}^p\) are the biases, \(n\) is the number of input units, \(p\) is the number of hidden units, \(f\) is a non-linear encoder function e.g. sigmoid.
After that the hidden representation \(h\) is mapped back to \(\tilde{x} \in \mathbb{R}^n \):
\[
\widehat{x} = g(\widehat{W}h + \widehat{b}),
\]
where \(\widehat{W} \in \mathbb{R}^{n \times p} \), \(b \in \mathbb{R}^n\), \(g\) is the identity function.
The weights and biases are found by gradient methods to minimize the cost function:
\[
J(W, b) = \frac{1}{N} \sum_{i=1}^{N}{\frac{1}{2} || \widehat{x}^{(i)} - x^{(i)} ||^2},
\]
where\(N\) is the number of inputs.

The overcompleted hidden layer is used to make the autoencoder extracting features.

Introducing spatial constraints with convolutions easily alignes the model of autoencoder to the convolutional autoencoder (CAE) and 3D convolutional autoencoder (3D-CAE).

In \cite{c4_sparse_encoder_Korea} authors added a sparsity constraint to prevent hidden layers of autoencoder from learning the identity function. They use 3D convolutions on the both sMRI and PET modalities and train the autoencoder on random \(5 \times 5 \times 5\) image patches. Max-pooling, fully-connected and softmax layers were applied after autoencoding. Mixing data of sMRI and PET modalities is performed at FC layer. The use of autoencoders allowed the authors using a subset of ADNI database to increase the classification accuracy by 4-6\% and leads to the level of 0.91 for AD/NC classification.

\begin{figure}[h]
\centering\includegraphics[width=1.0\linewidth]{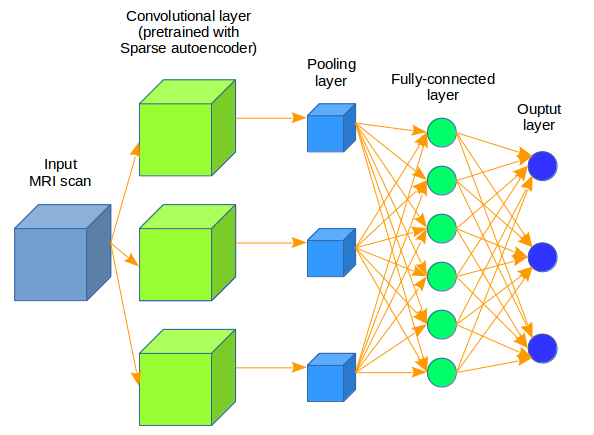}
\caption{Typical CNN architecture with CAE pretraining.}
\label{fig:cnn_1}
\end{figure}

Nearly the same approach with a sparse 3D autoencoder was used in \cite{c8_medical_UK} to classify sMRI images into 3 categories (AD/MCI/NC). The proposed network architecture is shown in Fig.\ref{fig:cnn_1}.  Larger obtained dataset selected from ADNI database and more accurate network parameters configuration allowed the authors to reach the accuracy of 0.954, 0.868 and 0.921 in AD/NC, AD/MCI and NC/MCI determination accordingly.

The authors of \cite{c2.2_3D_CNN_USA} extended the idea of applying autoencoders.  They proposed using three stacked 3D convolutional autoencoders instead of only one. Two fully-connected layers before the softmax were used for a progressive dimension reduction. The usage of stacked 3D CAE allowed the authors to achieve one of the best accuracy levels on 2265 images from ADNI database: 0.993, 1, 0.942 for AD/NC, AD/MCI and MCI/NC classification using sMRI images only. 

\subsubsection{Transfer Learning}

\begin{figure}[h]
\centering\includegraphics[width=1.0\linewidth]{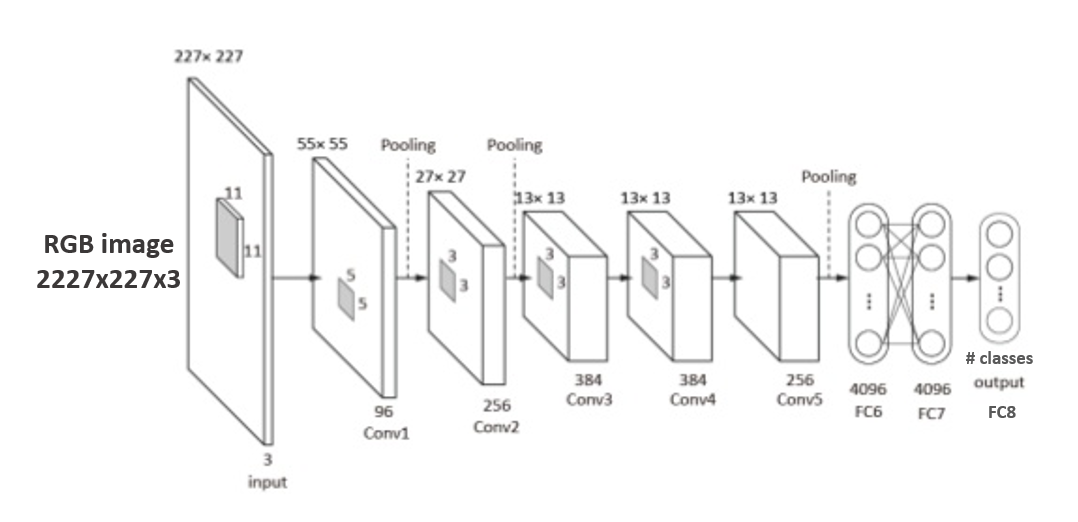}
\caption{AlexNet architecture. It includes 5 convolutional layers and 3 fully-connected layers.}
\label{fig:alexnet}
\end{figure}

Transfer learning is considered as the transfer of knowledge from one learned task to a new task in machine learning. In the context of neural networks, it is transferring learned parameters of a pretrained network to a new problem. Glozman and Liba in \cite{c1_report_Standford} used the widely known AlexNet \cite{alexnet}, pretrained on the ImageNet benchmark and fine-tuned the last 3 fully-connected layers (Fig.\ref{fig:alexnet}). The main problem of transfer learning is the necessity to transform the available data so that it corresponds to the network input. In \cite{c1_report_Standford} the authors created several 3-channel 2D images from the 3D input of sMRI and PET images by choosing central and nearby slices from axial, coronal and saggital projections. They  then interpolated the slices to the size \(227 \times 227\) compatible with AlexNet. Naturally one network was used for each projection. To augment the source data only mirror flipping was applied. This transfer learning based approach allowed the authors to reach 0.665 and 0.488 accuracy on 2-way (AD/NC) and 3-way (AD/MCI/NC) classifications accordingly on a subset of ADNI database.

In \cite{ex_3} authors apart from using the transfer learning technique proposed a convolutional neural network by involving Tucker tensor decomposition for classification of MCI subjects. The achieved accuracy on a subset of ADNI database containing 629 subjects was of 0.906.

\subsubsection{2D convolutional neural networks}

In \cite{c7.1_DeepAd_Canada}, \cite{c7.3_DeeapAd_Canada}, \cite{c7.2_DeeapAd_Canada} the authors compared the classification of structural and functional MRI images using one of the lightest  Deep architectures, the LeNet-5 architecture. They  transformed  the source 3D and 4D (in the case of fMRI) data to a batch of 2D images. LeNet-5 consists of two convolutional and two fully-connected layers. The reached level of accuracy for 2-class classification (AD/NC) was 0.988 for sMRI and 0.999 for fMRI images.

Billones et al. proposed in \cite{c5_demnet_Philippines} to use a modified 16-layered VGG network \cite{VGG} to classify sMRI images. The key feature of this paper was in using a 2D convolutional network to classify each slice of source data separately. The authors selected 20 central slices for each scan and the final score was calculated as the output of the last softmax layer of the network. The accuracy of each slice among all images was also studied, 17 slices were selected as representative, 3 slices (the first and two last slices in the image sequence) demonstrated lower level of accuracy. All in all authors reached a very good accuracy level: 0.983, 0.939, 0.917 for AD/NC, AD/MCI and MCI/NC classification using 900 sMRI images from a subset of ADNI database.

In \cite{Aderghal} Aderghal et al. used 3 central slices in each projection of a hippocampal ROI. The network architecture represented three 2D convolutional networks (one network per projection) that were joined in the last fully-connected layer. The reached accuracy for AD/NC, AD/MCI and MCI/NC classification is 0.914, 0.695 and 0.656 accordingly on a subset of ADNI database was nevertheless obtained not with siamese networks but by majority voting mechanism.

Ortiz-Suárez in \cite{r_1} explored the brain regions most contributing to Alzheimer's Disease by applying 2D convolutional neural networks to 2D sMRI brain images (coronal, sagittal and axial cuts). Using the dataset of 85 subjects the authors build a shallow 2D convolutional neural network. Then they create brain models for each filter at the CNN first layer and identify the filters with greatest discriminating power, thus choosing the most contributing brain regions. The authors demonstrated the largest differentiation between patients in the frontal pole region, which is known to host intellectual deficits related to the disease.

\subsubsection{Other networks}

\begin{figure}[h]
\centering\includegraphics[width=0.4\linewidth]{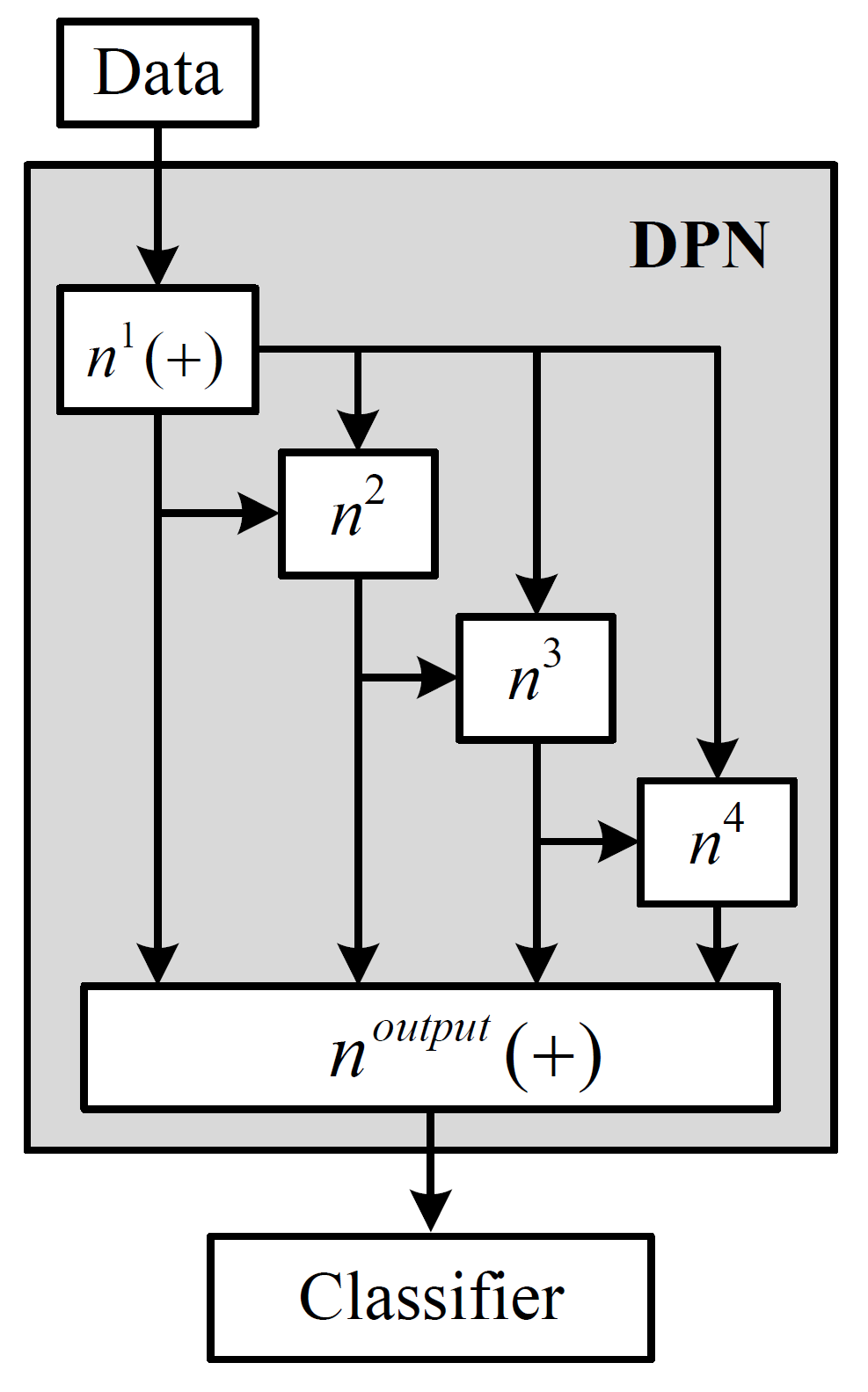}
\caption{An example of a DPN module.}
\label{fig:dpn}
\end{figure}

A new approach was proposed in \cite{c9_DPN_China}. Shi et al. used a deep polynomial network to analyze sMRI and PET images. It differs from classical CNNs by non-linearity of operations.  The building block of the architecture is shown in Fig.\ref{fig:dpn}. Here, \(n^i\) represents a layer of nodes, \((+)\) means a layer of nodes that calculate the weighted sum \(n(z) = \sum_{i}{w_i z_i}\), all other nodes compute \(n(z_1, z_2) = \sum_{i}{w_i (z_1)_i} (z_2)_i\). These blocks were combined into a deep network, the input layers were fed with the average intensity of the 93 ROIs selected on sMRI and PET brain images. A scheme  of the Deep Polynomial Network module is given in figure \ref{fig:dpn} below.

\begin{table*}[t]
\centering
\begin{footnotesize}
\begin{tabular}{|l|l|l|l|l|r|r|r|}
\hline
\multicolumn{1}{|c|}{\multirow{2}{*}{Algorithm}} & \multicolumn{1}{c|}{\multirow{2}{*}{Methodology}} & \multicolumn{1}{c|}{\multirow{2}{*}{Modalities}} & \multicolumn{1}{c|}{\multirow{2}{*}{Content}} & \multicolumn{1}{c|}{\multirow{2}{*}{Data (size)}} & \multicolumn{3}{c|}{Accuracy}                                                          \\ \cline{6-8} 
\multicolumn{1}{|c|}{} & \multicolumn{1}{c|}{} & \multicolumn{1}{c|}{} & \multicolumn{1}{c|}{} & \multicolumn{1}{c|}{} & \multicolumn{1}{c|}{AD/NC} & \multicolumn{1}{c|}{AD/MCI} & \multicolumn{1}{c|}{MCI/NC} \\ \hline

Magnin et al. \cite{volumetric:OBM} & Volumetric & sMRI & Full brain & custom (38) & 0.945 & - & - \\ \hline

\citeauthor{f3_Jenny_pcc} \cite{f3_Jenny_pcc}  & Feature-based & sMRI & 2 ROIs & ADNI (509) & 0.838 & 0.695 & 0.621 \\ \hline
Ebadi et al. \cite{g1_graph_ensemble_2017} & Feature-based & DTI & Full brain & custom (34) & 0.8 & 0.833 & 0.7 \\ \hline
Lee et al. \cite{g3_svm_DTI_Korea} & Feature-based & DTI & Full brain & LONI (141) & 0.977 & 0.977 & - \\ \hline
Lei et al. \cite{g2_sift_China} & Feature-based & sMRI + PET & Full brain & ADNI (398) & 0.969 & - & 0.866 \\ \hline
\citeauthor{f5_Jenny} \cite{f5_Jenny} & Feature-based & sMRI + DTI & 1 ROI & ADNI (203) & 0.902 & 0.766 & 0.794 \\ \hline
Salvatore et al. \cite{salvatore2015magnetic} & Feature-based & sMRI & Full brain & ADNI (509) & 0.76 & - & - \\ \hline
\citeauthor{landmark_AD} \cite{landmark_AD} & Feature-based & sMRI & Full brain & ADNI (428) & 0.837 & - & - \\ \hline

Vu et al. \cite{c4_sparse_encoder_Korea} & NN-based & sMRI + PET & Full brain & ADNI (203) & 0.91 & - & - \\ \hline
Payan and Montana \cite{c8_medical_UK} & NN-based & sMRI & Full brain & ADNI (2265) & 0.993 & \textbf{1.0} & \textbf{0.942} \\ \hline
Glozman and Liba \cite{c1_report_Standford} & NN-based & sMRI + PET & Full brain & ADNI (1370) & 0.665 & - & - \\ \hline
Sarraf et al. \cite{c7.1_DeepAd_Canada} & NN-based & sMRI, fMRI & Full brain & ADNI (302) & \textbf{0.988, 0.999} & - & - \\ \hline
Billones et al. \cite{c5_demnet_Philippines} & NN-based & sMRI & Full brain & ADNI (900) & 0.983 & 0.939 & 0.917 \\ \hline
Aderghal et al. \cite{Aderghal} & NN-based & sMRI & 1 ROI & ADNI (815) & 0.914 & 0.695 & 0.656 \\ \hline
Shi et al. \cite{c9_DPN_China} & NN-based & sMRI + PET & Full brain & ADNI (202) & 0.971 & - & 0.872 \\ \hline
Korolev et al. \cite{c3_Skolkovo} & NN-based & sMRI & Full brain & ADNI (231) & 0.79-0.8 & - & - \\ \hline
Suk et al. \cite{c6_deep_ensemble_sparse_regr_Korea} & NN-based & sMRI & 93 ROIs & ADNI (805) & 0.903 & - & 0.742 \\ \hline
Luo et al. \cite{ex_2} & NN-based & sMRI & 7 ROIs & ADNI (81) & 0.83 & - & - \\ \hline
Wang et al. \cite{ex_3} & NN-based & sMRI & Full brain & ADNI (629) & - & - & 0.906 \\ \hline
Li et al. \cite{ex_5} & NN-based & sMRI & 1 ROI & ADNI (1776) & 0.965 & 0.67 & 0.622 \\ \hline
Cheng et al. \cite{ex_7} & NN-based & sMRI & 27 ROIs & ADNI (1428) & 0.872 & - & - \\ \hline
Li et al. \cite{ex_9} & NN-based & sMRI & Full brain & ADNI (832) & 0.91 & 0.877 & 0.855 \\ \hline
Liu et al. \cite{liu2018multi} & NN-based & sMRI + PET & Full brain & ADNI (397) & 0.93 & - & - \\ \hline

\end{tabular}
\end{footnotesize}
\caption{Comparison of different state-of-the-art classification methods of Alzheimer Disease diagnostics.}
\label{table:comparison}
\end{table*}

This architecture allowed the authors to reach very good level of accuracy: 0.971, 0.872 for AD/NC, MCI/NC classification. The used algorithm also demonstrated a good level of accuracy (0.789) for MCI-C/MCI-NC determination, where MCI-C  stands for MCI patients that lately converted to AD and MCI-NC stands for MCI patient that were not converted.

In \cite{c3_Skolkovo} the authors compared the residual (ResNet) and plain 3D convolutional neural networks for sMRI image classification. Here the authors examined the four binary classification tasks  AD/LMCI/EMCI/NC, where LMCI and EMCI stands for the late and early MCI stages accordingly. Both networks demonstrated nearly the same performance level, the best figures being obtained for AD/NC classification with  0.79-0.8 accuracies, using 231 sMRI images from a subset of ADNI database.

Residual convolutional networks having shown good performances in computer vision tasks, Li et al. in \cite{ex_5} have also proposed a deep network with residual blocks to preform ordinal ranking.  They compared their model to classical multi-category classification techniques. Data of the only one hippocampal ROI from 1776 sMRI images of ADNI database were used. The final accuracy performance of the proposed method is 0.965, 0.67 and 0.622 for AD/NC, AD/MCI and MCI/NC classification accordingly.

A so-called spectral convolutional neural network was proposed in \cite{ex_9}. It combines classical convolutions with the ability to learn some topological brain features. \citeauthor{ex_9} represented a subject's brain as a graph with a set of ROIs as nodes and edges computed using Pearson correlation from a brain grey matter. With a subset of ADNI database containing sMRI images of 832 subjects the authors achieved the classification accuracy 0.91, 0.877, 0.855 for AD/NC, AD/MCI and MCI/NC classification.

In \cite{c6_deep_ensemble_sparse_regr_Korea} \citeauthor{c6_deep_ensemble_sparse_regr_Korea} tried to combine two different methods: sparse regression and convolutional neural networks. The authors got different sparse representations of the 93 ROIs of the sMRI data by varying the sparse control parameter, which allowed them to produce different sets of selected features. Each representation is a vector, so the result of generating multiple representations can be treated as a matrix. This matrix is then fed to the convolutional neural network with 2 convolutional layers and 2 fully-connected layers. This approach led to the classification accuracy level of 0.903 and 0.742 for AD/NC and MCI/NC classification.

Although, the research in the area of Alzheimer's Disease diagnostics with deep neural networks is very extensive, the general trend can be identified: 2D CNNs are being progressively replaced by 3D CNNs. Furthermore, data fusion from different modalities is surely a way to follow. Hence in the next section we propose our new architecture of 3D CNN with data fusion.

\section{Proposed network architecture}
\label{sec:network}

In this work we propose a new 3D Inception-based convolutional neural network architecture, based on the idea of improved utilization of the computer resources inside the network, first mentioned in \cite{inception} for 2D case.

\begin{figure}[h]
\centering\includegraphics[width=\linewidth]{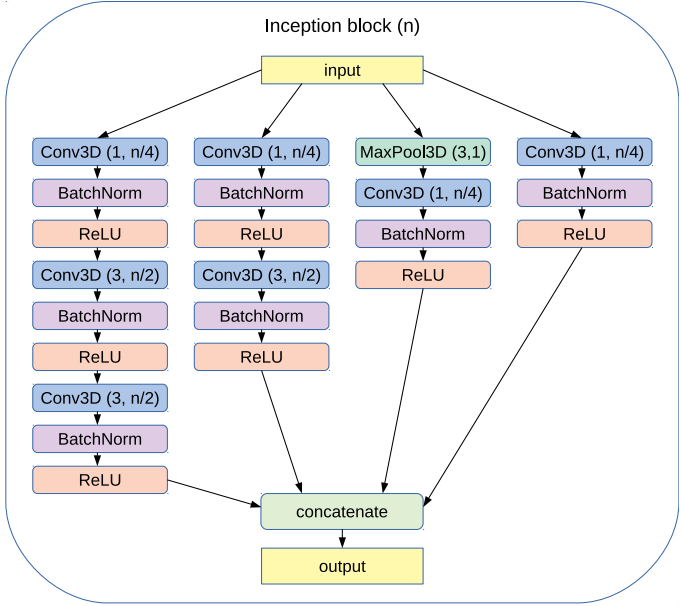}
\caption{Inception block of the proposed network architecture. Here $n$ stands for the number of features in the input of the block, Conv3D$(s, m)$ stands for 3D convolution with $m$ filters of $s \times s \times s$ size, MaxPool3D$(p, q)$ stands for 3D max-pool operation with pool size $p$ and stride $q$.}
\label{fig:inception_block}
\end{figure}

The main building block of the network is an Inception block (Fig. \ref{fig:inception_block}).  To eliminate the need of choosing the specific layer type at each level of the network Inception block uses 4 different bands of layers simultaneously. Besides that, a number of $1 \times 1 \times 1$ convolution filters are used to significantly reduce the number of network parameters by decreasing the dimension of the feature space. In particular, the first band of the block performs a two successive $3 \times 3 \times 3$ convolutions (equivalent to $5 \times 5 \times 5$ filter), the second band performs one $3 \times 3 \times 3$ convolution, third band performs a max-pool operation, fourth band performs $1 \times 1 \times 1$ convolution. Besides that, first three bands use $1 \times 1 \times 1$ convolution at the beginning. Each convolution layer is followed with batch normalization layer \cite{batch_norm_2015} and a ReLU. The number of features in each convolution depends on the input and is shown in Fig.\ref{fig:inception_block}. Thus, the output of the Inception block increases the feature dimension of data in 1.5 times compared to its input. All these tricks  substantially reduce the number of parameters inside the network, while at the same time batch normalization layers accelerate network training.

\begin{figure}[h]
\centering\includegraphics[width=0.3\linewidth]{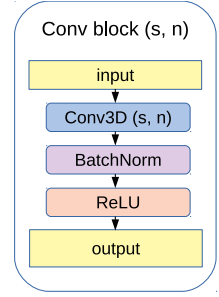}
\caption{Conv block of the network. Here Conv3D$(s, n)$ stands for 3D convolution with $n$ filters of $s \times s \times s$ size.}
\label{fig:conv_block}
\end{figure}

\begin{figure*}[h]
\centering\includegraphics[width=0.9\linewidth]{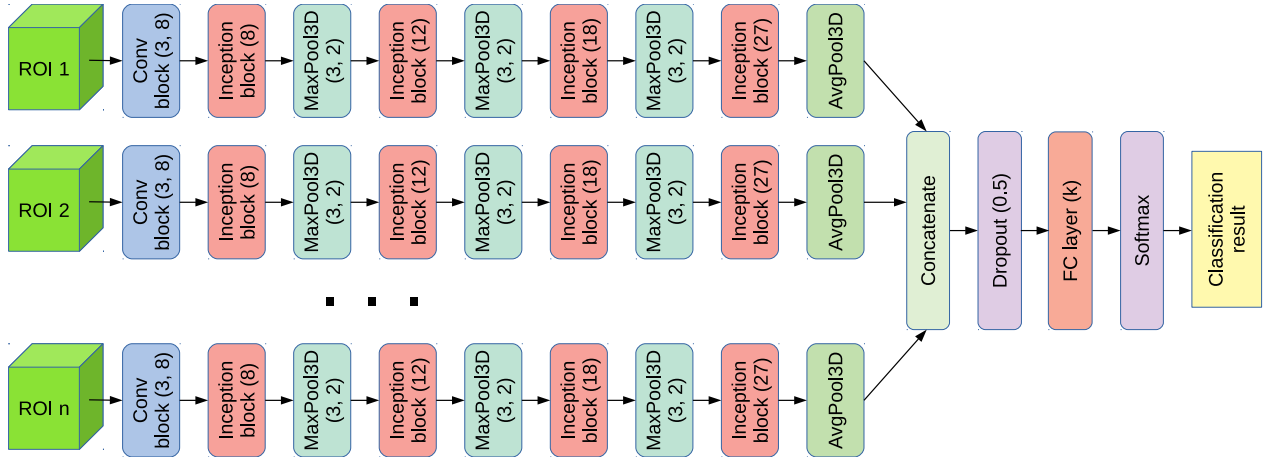}
\caption{Proposed network architecture. Here MaxPool3D$(p, q)$ stands for 3D max-pool operation with pool size $p$ and stride $q$, FC layer $(k)$ stands for the fully-connected layer with $k$ outputs, the structure of Conv block$(s, n)$ is presented in Fig.\ref{fig:conv_block}.}
\label{fig:inception_net}
\end{figure*}

A preliminary $3 \times 3 \times 3$ Conv block with the 4 sequent combinations of Inception block with 3D max-pooling layer form a pipeline of the proposed network architecture. This pipeline transforms the source spatial data to the feature space. The last modification to reduce the number of network parameters compared to the conventional AlexNet-like networks \cite{alexnet} is to place a 3D average-pooling layer at the end of the pipeline instead of the fully-connected layer. For each ROI in the brain scan and for each modality we use a separate described above pipeline. Finally, all pipelines are concatenated and with the following dropout, fully-connected and softmax layers produce the classification result (Fig. \ref{fig:inception_net}). Thus the described network is a siamese network which performs the late fusion of the data from input ROIs.

The usage of batch normalization as mentioned earlier allows us to speed up the network training process and according to \cite{batch_norm_2015} eliminate the necessity of using the pretraining techniques (e.g. autoencoders). Batch normalization partially plays a role of regularization as it allows each layer of a network to be trained less dependent on other layers \cite{batch_norm_2015}.

\section{Experiments and results}
\label{sec:Experiments_and_results}

In this work we compare the proposed network design (Fig.\ref{fig:inception_net}) with the conventional AlexNet-based network (Fig.\ref{fig:conv_net}) for the Alzheimer's Disease detection. We also analyze the dependency of used image modalities (sMRI, DTI) on the classification results. The obtained results are shown in Table \ref{table:results} and are discussed in Section \ref{sec:Results}. 

\begin{table}[]
\centering
\begin{tabular}{lccc}
\hline
\multicolumn{1}{c}{}                                                           	& AD	& MCI	& NC  \\ \hline
Subjects                                                                       	& 53	& 228	& 250  \\
Subjects in train set                                                             	& 35	& 192	& 212  \\
Subjects in validation set                                                             	& 3		& 21	& 23  \\
Subjects in test set                                                             	& 15	& 15	& 15  \\ \hline
\end{tabular}
\caption{Number of subjects for each class.}
\label{table:samples}
\end{table}

\subsection{Data selection}
Data  used  in  the  preparation  of  this  article  were  obtained  from  the  Alzheimer’s  Disease Neuroimaging  Initiative  (ADNI)  database  (\url{http://adni.loni.usc.edu}).
The  ADNI  was  launched  in 2003 as  a  public-private  partnership,  led  by  Principal  Investigator  Michael  W. Weiner, MD. The primary goal of ADNI has been to test whether serial magnetic resonance imaging (MRI),  positron  emission  tomography  (PET),  other  biological  markers,  and  clinical  and neuropsychological  assessment  can  be  combined  to  measure  the  progression  of  mild cognitive impairment (MCI) and early Alzheimer’s Disease (AD).
For up-to-date information, see \url{www.adni-info.org}.
We selected 531 subjects: 53 AD, 228 MCI and 250 NC patients from ADNI 2-Go and ADNI 3 datasets (Table \ref{table:samples}). For each patient there is a T1-weighted sMRI image as well as a DTI image. Table \ref{table:demographic} presents a summary of the demographic characteristics of the selected subjects including the age, gender and Mind Mental State Examination (MMSE) score of cognitive functions. In our case, the number of images in the dataset is limited by the availability of DTI data. We focus on the hippocampal ROI and surrounding region in the brain scans. 

\begin{table}[]
\centering
\begin{footnotesize}
\begin{tabular}{cccccc}
Dataset & Diagnosis & Subjects & Age & \begin{tabular}[c]{@{}c@{}}Gender\\ (F/M)\end{tabular} & MMSE \\ \hline
\multirow{3}{*}{\begin{tabular}[c]{@{}c@{}}ADNI\\ 2-Go\end{tabular}} & AD & 48 & \begin{tabular}[c]{@{}c@{}}[55.72, 91.53]\\ 75.65 $\pm$ 8.63\end{tabular} & 20/28 & \begin{tabular}[c]{@{}c@{}}23.0\\ $\pm$ 2.42\end{tabular} \\
 & MCI & 108 & \begin{tabular}[c]{@{}c@{}}[55.32, 91.88]\\ 73.46 $\pm$ 7.47\end{tabular} & 42/66 & \begin{tabular}[c]{@{}c@{}}27.39\\ $\pm$ 1.99\end{tabular} \\
 & NC & 58 & \begin{tabular}[c]{@{}c@{}}[55.32, 91.88]\\ 73.46 $\pm$ 7.47\end{tabular} & 30/28 & \begin{tabular}[c]{@{}c@{}}28.88\\ $\pm$ 1.99\end{tabular} \\ \hline
\multirow{3}{*}{\begin{tabular}[c]{@{}c@{}}ADNI\\ 3\end{tabular}} & AD & 5 & \begin{tabular}[c]{@{}c@{}}[55.97, 86.09]\\ 71.62 $\pm$ 11.49\end{tabular} & 2/3 & - \\
 & MCI & 120 & \begin{tabular}[c]{@{}c@{}}[57.72, 95.93]\\ 75.79 $\pm$ 7.82\end{tabular} & 55/65 & - \\
 & NC & 192 & \begin{tabular}[c]{@{}c@{}}[55.79, 95.38]\\ 74.93 $\pm$ 8.00\end{tabular} & 116/76 & - \\ \hline
\end{tabular}
\end{footnotesize}
\caption{Demographic description of the ADNI group, values are denoted as
intervals and as mean $\pm$ std.}
\label{table:demographic}
\end{table}

A preprocessing is performed on all used DTI brain images. It includes correction of eddy currents and head motion, skull stripping with Brain Extraction Tool (BET) \cite{BET} and fitting of diffusion tensors to the data with DTI-fit module of the FSL software library \cite{FSL}. Fitting step generates MD and FA maps. In the current work we focus only on MD maps of DTI images. To use a normalized anatomical atlas for ROI selection the MD images are affinely co-registered to corresponding sMRI scans. After such co-registration both image modalities are spatially normalized onto the Montreal Neurological Institute (MNI) brain template \cite{MNI}.
Thus, after the preprocessing step, for each patient there is a sMRI and MD-DTI aligned images of the same resolution of \(121 \times 145 \times 121\) voxels.

For the further analysis on each image we select two ROIs (left and right lobes of the hippocampus) as the most discriminative parts of human brain for Alzheimer's Disease analysis \cite{pierrick}. The ROI selection is performed using atlas AAL \cite{AAL}, the resolution of both hippocampal areas is \(29 \times 29 \times 29 \times 29\) voxels.

\subsection{Dataset formation and augmentation}

We divide the used image dataset into train, validation and test sets. For test set we randomly select 15 subjects from each class, all the remaining subjects are split randomly into validation and train sets with 9:1 ratio for each class (Table \ref{table:samples}).

The common problem of using limited dataset for training a neural network is overfitting. To enlarge the amount of data and prevent the network overfitting we perform data augmentation. The augmentation process is performed for the train set only. Besides that, as in many other medical problems the used dataset is imbalanced: the number of patients with AD is almost 5 times smaller than the number of patients with MCI or NC. To eliminate the effect of different class capacities on the network training process we perform a balancing procedure during training. Improvements of classification results in the case of using data balancing procedure was demonstrated in \cite{r_1}. The main distinctive feature of the method we use is that class balancing is performed together with data augmentation on the fly.

The augmentation process is described with a parameter \(\tau\), which controls the level of augmentation (i.e. the amount of new images generated from the source one). All new images are generated from the source images by random shift up to $2$ pixels in each of the three dimensions $XYZ$.This is a typical domain augmentation technique in brain image analysis. Indeed, the shifts compensate for imprecision of alignment of individual brains on the MNI template. 

A batch of size $\eta$ of training data is formed as follows. A random class (from the used ones) is selected, then a random image of this class is chosen from the dataset and finally this image is randomly augmented within the boundary augmentation parameters. This sequence of operations is performed $\eta$ times, thus forming the training batch.

The number of images used in one training epoch is chosen as the number of subjects in train set multiplied by augmentation factor $\tau$. In this work we have chosen $\tau=5$. Thus with $439$ subjects in train set this leads to $2195$ resulting images the network is trained on during one epoch. In the case of the batch size $\eta=15$ the number of training iterations for one epoch is $146$.
 
To prevent the network overfitting the random split of the subjects in train and validation sets is repeated after each training epoch. This approach leads to the effective usage of the available training data.

\subsection{Implementation details}
\label{sec:Implementation}

\begin{figure*}[h]
\centering\includegraphics[width=0.9\linewidth]{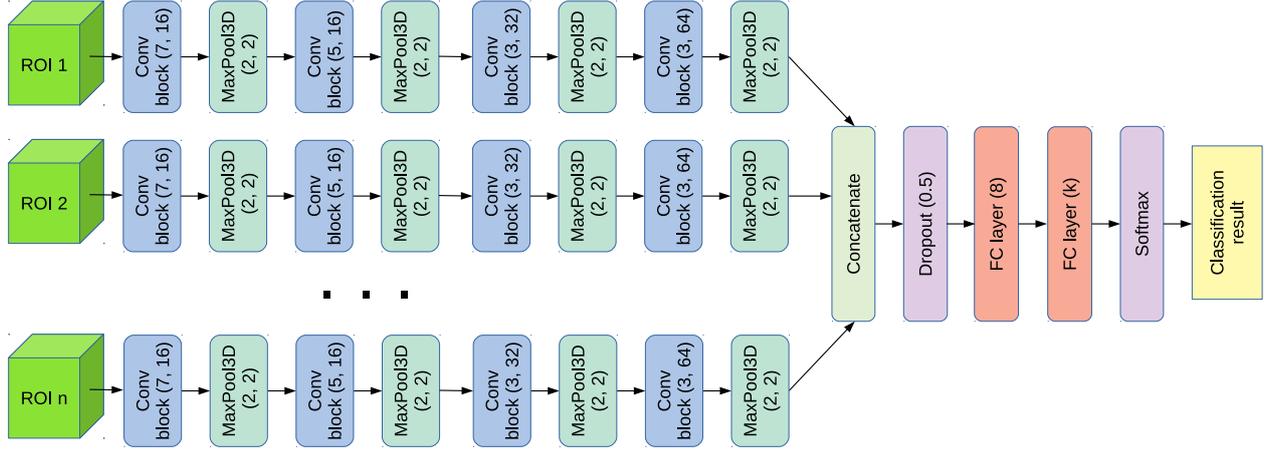}
\caption{AlexNet-based network used for comparison. Here MaxPool3D$(p, q)$ stands for 3D max-pool operation with pool size $p$ and stride $q$, FC layer $(k)$ stands for the fully-connected layer with $k$ outputs, Conv block$(s, n)$ is presented in Fig.\ref{fig:conv_block}.}
\label{fig:conv_net}
\end{figure*}

To analyze the efficiency of the designed network described in Section \ref{sec:network} we compare it to the conventional AlexNet-based network with the same depth and similar structure. The architecture of the used AlexNet-based is shown in Fig.\ref{fig:conv_net}. Both networks contain 4 convolutional blocks (Inception blocks in the case of the proposed network), each followed with 3D max-pooling layer, the number of input features for Inception and Conv blocks are shown in Figs.\ref{fig:inception_net},\ref{fig:conv_net}.

As it was discussed earlier for each ROI and each modality we perform a base pipeline and then do a late fusion. To compare sMRI and DTI modalities and analyze their applicability to Alzheimer's Disease detection we consider neural networks with the following pipeline inputs (the corresponding abbreviations used below are given at the beginning of the lines):
\begin{enumerate}
\item \textit{DTI\_L+DTI\_R}: left hippocampus on MD-DTI and right hippocampus on MD-DTI images
\item \textit{sMRI\_L+sMRI\_R}: left hippocampus on sMRI and right hippocampus on sMRI images
\item \textit{sMRI\_L+sMRI\_R+DTI\_L+DTI\_R}: left hippocampus on sMRI, right hippocampus on sMRI, left hippocampus on MD-DTI and right hippocampus on MD-DTI images
\end{enumerate}

Thus, if using 4 ROIs (\textit{sMRI\_L+sMRI\_R+DTI\_L+DTI\_R}) and the described above configurations (Figs.\ref{fig:inception_net}, \ref{fig:conv_net}) a total number of parameters of the network in the case of proposed architecture is 71,939 compared to the 431,075 in the case of the AlexNet-based network.

A uniform Xavier initialization \cite{xavier} of weights are applied for both networks. A categorical cross-entropy loss is used as the loss function  \cite{goodfellow2016deep}. The networks are trained with the same dataset using RMSprop optimizer \cite{RMSprop}, dividing the gradient by a running average of its current magnitude. During the training process the learning rate automatically decreases when the test loss falls on the plateau and stops decreasing. The initial learning rate value is manually tuned depending on the batch size and the network type. In particular, if the batch size is increased in $k$ times the initial learning rate is decreased in $k$ times accordingly \cite{minibatch_lr}, while the initial value of the learning rate is chosen manually for each network architecture.

Both networks were implemented using open source neural network library Keras \cite{keras} with TensorFlow \cite{tensorflow} backend. The experiments were performed on two configurations: a personal computer with Intel(R) Core(R) i7-7700HQ CPU and Nvidia GeForce GTX 1070 GPU and free Google Colaboratory cloud server with Nvidia Tesla K80 GPU.

\subsection{Results}
\label{sec:Results}

\begin{table}[]
\centering
\begin{tabular}{llccc}
\textbf{Task} & \multicolumn{1}{c}{\textbf{\begin{tabular}[c]{@{}c@{}}Network\\ type\end{tabular}}} & \textbf{\begin{tabular}[c]{@{}c@{}}Accuracy\\ {[}95\% CI{]}\end{tabular}} & \textbf{\begin{tabular}[c]{@{}c@{}}Sensitivity\\ {[}95\% CI{]}\end{tabular}} & \textbf{\begin{tabular}[c]{@{}c@{}}Specificity\\ {[}95\% CI{]}\end{tabular}} \\ \hline
\multicolumn{1}{c}{\multirow{2}{*}{\begin{tabular}[c]{@{}c@{}}AD/\\ MCI/\\ NC\end{tabular}}} & \begin{tabular}[c]{@{}l@{}}AlexNet-\\ based\end{tabular} & \begin{tabular}[c]{@{}c@{}}0.62\\$\pm$0.142\end{tabular} & \begin{tabular}[c]{@{}c@{}}-\end{tabular} & \begin{tabular}[c]{@{}c@{}}-\end{tabular} \\
\multicolumn{1}{c}{} & proposed & \begin{tabular}[c]{@{}c@{}}\textbf{0.689}\\ $\pm$0.135\end{tabular} & \begin{tabular}[c]{@{}c@{}}-\end{tabular} & \begin{tabular}[c]{@{}c@{}}-\end{tabular} \\ \hline
\multirow{2}{*}{\begin{tabular}[c]{@{}l@{}}AD/\\ NC\end{tabular}} & \begin{tabular}[c]{@{}l@{}}AlexNet-\\ based\end{tabular} & \begin{tabular}[c]{@{}c@{}}0.9\\$\pm$0.107\end{tabular} & \begin{tabular}[c]{@{}c@{}}0.867\\$\pm$0.121\end{tabular} & \begin{tabular}[c]{@{}c@{}}0.933\\$\pm$0.089\end{tabular} \\
 & proposed & \begin{tabular}[c]{@{}c@{}}\textbf{0.933}\\ $\pm$0.089\end{tabular} & \begin{tabular}[c]{@{}c@{}}\textbf{0.933}\\ $\pm$0.089\end{tabular} & \begin{tabular}[c]{@{}c@{}}\textbf{0.933}\\ $\pm$0.89\end{tabular} \\ \hline
\multirow{2}{*}{\begin{tabular}[c]{@{}l@{}}AD/\\ MCI\end{tabular}} & \begin{tabular}[c]{@{}l@{}}AlexNet-\\ based\end{tabular} & \begin{tabular}[c]{@{}c@{}}0.833\\$\pm$0.133\end{tabular} & \begin{tabular}[c]{@{}c@{}}0.8\\$\pm$0.143\end{tabular} & \begin{tabular}[c]{@{}c@{}}0.867\\$\pm$0.122\end{tabular} \\
 & proposed & \begin{tabular}[c]{@{}c@{}}\textbf{0.867}\\ $\pm$0.122\end{tabular} & \begin{tabular}[c]{@{}c@{}}\textbf{0.8}\\ $\pm$0.143\end{tabular} & \begin{tabular}[c]{@{}c@{}}\textbf{0.933}\\ $\pm$0.089\end{tabular} \\ \hline
\multirow{2}{*}{\begin{tabular}[c]{@{}l@{}}MCI/\\ NC\end{tabular}} & \begin{tabular}[c]{@{}l@{}}AlexNet-\\ based\end{tabular} & \begin{tabular}[c]{@{}c@{}}0.667\\$\pm$0.169\end{tabular} & \begin{tabular}[c]{@{}c@{}}0.8\\$\pm$0.143\end{tabular} & \begin{tabular}[c]{@{}c@{}}0.53\\$\pm$0.179\end{tabular} \\
 & proposed & \begin{tabular}[c]{@{}c@{}}\textbf{0.733}\\ $\pm$0.158\end{tabular} & \begin{tabular}[c]{@{}c@{}}\textbf{0.8}\\ $\pm$0.143\end{tabular} & \begin{tabular}[c]{@{}c@{}}\textbf{0.86}\\ $\pm$0.122\end{tabular} \\ \hline
\end{tabular}
\caption{Classification results on the test set using one ROI left and right and two modalities  (\textit{sMRI\_L+sMRI\_R+DTI\_L+DTI\_R}). Here \(\alpha \pm \beta\) denotes the value of metric and it's 95\% confidence interval. \(\alpha\) is a metric value, \( [\alpha - \beta, \alpha + \beta] \) is the confidence interval.}
\label{table:results}
\end{table}

In this work we train and evaluate one ternary AD/MCI/NC classifier and 3 binary AD/NC, AD/MCI, MCI/NC classifiers to detect the corresponding classes. The obtained results are shown in Table \ref{table:results}.

\begin{figure}[h]
\centering\includegraphics[width=\linewidth]{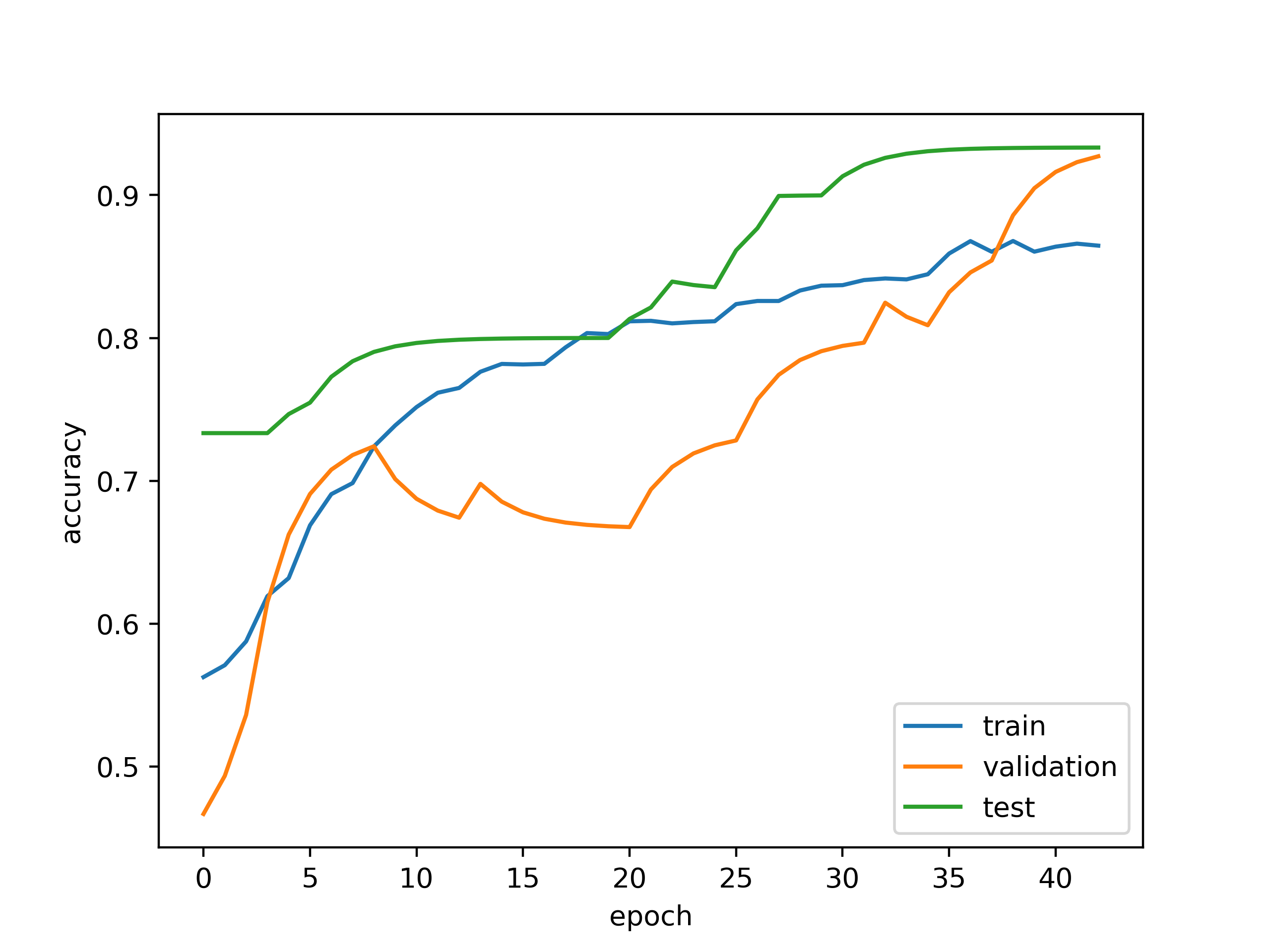}
\caption{Accuracy plot for binary AD/NC classification with the proposed network for train, validation and test sets.}
\label{fig:accuracy_plot}
\end{figure}

\begin{figure}[h]
\centering\includegraphics[width=\linewidth]{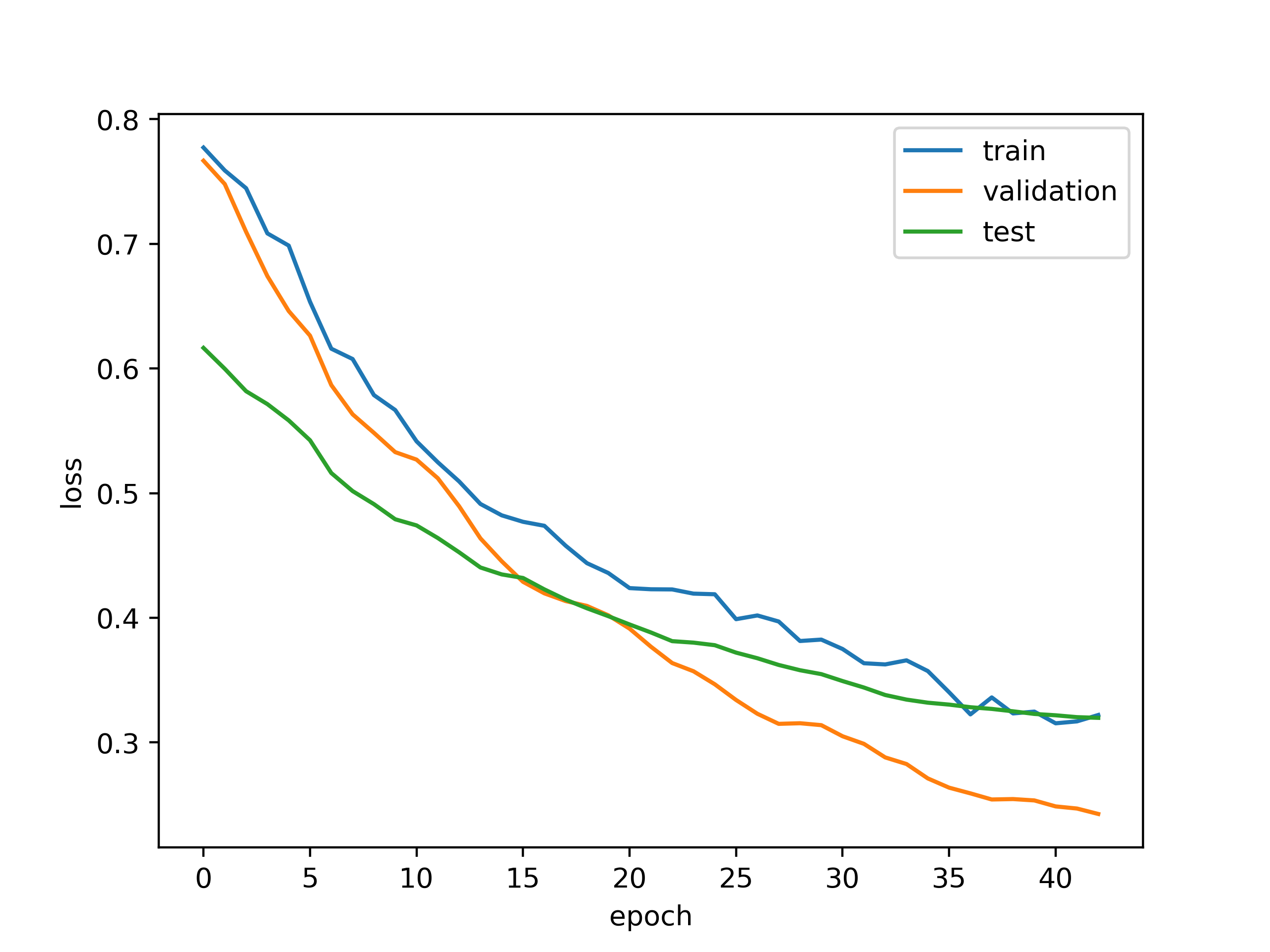}
\caption{Loss plot for binary AD/NC classification with the proposed network for train, validation and test sets.}
\label{fig:loss_plot}
\end{figure}

Here to evaluate and score each experiment we use accuracy as a reference metric. Along with accuracy (ACC) we also report the values of sensitivity (SEN) and specificity (SPC). We should also notice the absence of commonly used balanced accuracy (BAC) metric. That is because we use an already balanced test set, as the number of subjects used for testing in each class is the same (Table \ref{table:samples}). So in the current case all reported accuracy values in this work are equal to the balanced accuracy values. To perform an interval estimation of the classification we also report a 95\% confidence interval for all described metrics using Wilson score interval \cite{wilson_1,wilson_2}. The confidence interval is calculated as:
\[
	CI = val \pm \theta \cdot \sqrt[]{\frac{val (1-val)}{n}},
\]
where \(\theta\) is a constant corresponding to confidence range (in the case of 95\% range \(\theta=1.96\)), \(n\) is a number of samples in the set and \(val\) is a value of metric for which we calculate a confidence interval. Here we should also notice, that the width of the confidence interval depends on the number of samples in the set. Thus, in our case (Table \ref{table:samples}) $n=30$ for binary AD/NC, AD/MCI, MCI/NC classifiers and $n=45$ for ternary AD/MCI/NC classifier.

\begin{table}[]
\centering
\begin{tabular}{clcc}
\textbf{Task} & \multicolumn{1}{c}{\textbf{\begin{tabular}[c]{@{}c@{}}Network\\ type\end{tabular}}} & \textbf{\begin{tabular}[c]{@{}c@{}}Convergence\\ on epoch\end{tabular}} & \textbf{\begin{tabular}[c]{@{}c@{}}Training\\ time {[}min{]}\end{tabular}} \\ \hline
\multirow{2}{*}{AD/MCI/NC} & AlexNet-based & 30 & 73 \\
 & proposed & \textbf{28} & \textbf{45} \\ \hline
\multirow{2}{*}{AD/NC} & AlexNet-based & \textbf{34} & 54 \\
 & proposed & 36 & \textbf{42} \\ \hline
\multirow{2}{*}{AD/MCI} & AlexNet-based & 26 & 40 \\
 & proposed & \textbf{21} & \textbf{22} \\ \hline
\multirow{2}{*}{MCI/NC} & AlexNet-based & 29 & 48 \\
 & proposed & \textbf{24} & \textbf{28}
\end{tabular}
\caption{Number of epochs at which the optimal model was obtained and the average time spent on training in minutes. The results were obtained for networks with sMRI and MD-DTI input ROIs and batch size = 15 using Nvidia GeForce GTX 1070 GPU.}
\label{table:epochs}
\end{table}

As an example, accuracy and loss plots for the case of binary AD/NC classification with the proposed network architecture are shown in Figs.\ref{fig:accuracy_plot},\ref{fig:loss_plot}. One can notice that the accuracy curve on the test set demonstrates better performance than on validation set. It can be explained by the larger size of the test set compared to the validation set and by the training strategy we deployed with random cross-validation reshuffling, which leads to better covering of the training set.

During the experiments it was found that in all classification cases (ternary AD/MCI/NC and binary AD/NC, AD/MCI, MCI/NC problems) the designed network demonstrates significantly better results than the conventional AlexNet-based network (Table \ref{table:results}). This can be explained with the fewer weights in the designed network while maintaining the same depth, less amount of data needed to train the network, faster and more robust training process as a result.

In this work we also analyze the influence of the used image modalities (MD-DTI, sMRI) as well as their fusion on the classification results. Thus, we compare 3 types of inputs that were mentioned in Section \ref{sec:Implementation}. For the proposed network architecture in all classification cases the results obtained with 4 input ROIs (left hippocampus on MD-DTI + right hippocampus on MD-DTI + left hippocampus on sMRI + right hippocampus on sMRI) are the best. The results obtained with left and right hippocampal ROIs on sMRI data are worser, the results obtained with left and right hippocampal ROIs on MD-DTI data are the worst of the three. Due to this reason only the case of using both sMRI and MD-DTI input ROIs is included in the result Table \ref{table:results}.

The number of epochs needed for convergence to the optimal model in the case of the proposed network and AlexNet-based one is demonstrated in Table \ref{table:epochs}.  Thus, the proposed network can be trained $\sim 1.7$ times faster compared to the AlexNet-based one.
The time required for testing one batch of data is nearly the same in both cases: 4 and 7 seconds for binary and ternary classification in the case of the AlexNet-based network and 5 and 8.5 seconds for binary and ternary classification in the case of the proposed network.

All in all we succeeded to achieve the classification accuracy of 0.933, 0.867 and 0.733 for binary AD/NC, AD/MCI and MCI/NC classification problems respectively and 0.689 for ternary AD/MCI/NC classification problem.

\section{Discussion}
\label{sec:Discussion}
To compare the performance of our method with the state-of-the-art we come back to the Table \ref{table:comparison}. The general observation is that relatively new feature-based and neural network-based methods demonstrate very good level of performance compared to the classical volumetric methods that are performed manually by medical experts.

It should be mentioned, that the direct comparison of our method with the reviewed algorithms for Alzheimer's Disease diagnostics is impossible as the results were obtained using images from several databases and with datasets of different size (see Table \ref{table:comparison}). Moreover, various classification problems were challenged: although most papers focus on the 3-class AD/MCI/NC classification, some of them consider only 2-class AD/NC classification \cite{c4_sparse_encoder_Korea, c7.1_DeepAd_Canada, c7.2_DeeapAd_Canada, c7.3_DeeapAd_Canada} and even 4-class AD/eMCI/lMCI/NC classification \cite{c3_Skolkovo}. Also \cite{c5_demnet_Philippines,c6_deep_ensemble_sparse_regr_Korea} deserve special attention as the authors try to solve a  problem of Alzheimer's converters prediction. 

Nevertheless, with the accuracies of 0.933, 0.867 and 0.733 for binary classification problems of AD/NC, AD/MCI and MCI/NC the proposed solution with 3D Inception-based CNN and fusion of modalities outperforms similar approaches such as \cite{ex_5} with ROI, but only on one modality despite much larger dataset used there for training.

Furthermore, compared with feature-based methods with fusion of the same sMRI and DTI modalities in \cite{f1_Jenny_MKL} the CNN classifiers confirm their better performance. As far as full-brain approaches are concerned, such as \cite{c7.1_DeepAd_Canada},\cite{ex_9}, the consensus cannot be obtained, as the best performances are shown for the work of Payan and Montana with quite a large dataset on a single sMRI modality \cite{c8_medical_UK}. Her we should also notice that full-brain schemes require much stronger computational resources as the full resolution 3D scans have to be submitted to the network architecture at once.

The proposed 3D Inception-based network architecture better utilizes the interior network resources and despite the seeming sophistication contains less weights compared to the AlexNet-based network with the same depth. In particular, the total number of parameters in the proposed network in case of 2 ROIs of sMRI and DTI modalities and 4 layers is 6 times less compared to the similar AlexNet-based network. This optimization of network architecture increases the accuracy by 3-6\% for binary classification problems and by almost 7\% in the case of the ternary classification (Table \ref{table:results}).

It should also be noticed that as the proposed architecture contains less weights than a similar AlexNet-based network, it can be trained with larger batch size on the same GPU. But at the same time, in the case of equal batch size the number of training epochs needed for the convergence to the optimal model is nearly the same for both proposed and AlexNet-based networks (Table \ref{table:epochs}).
Furthermore, according to Table \ref{table:epochs} due its more "light" structure the proposed network requires  $\sim 1.7$ less time to be trained, which could be also a practical advantage in the case of expanding the training dataset and fine-tuning the network.


\section{Conclusion and perspectives}
\label{sec:Conclusion}
In this paper we have proposed a new design of a multimodal 3D CNN for Alzheimer's Disease diagnostics inspired by an Inception model which has proven efficient for general purpose 2D image databases. The proposed network is constructed with the emphasis on the interior resource utilization. It contains less weights comparing to the conventional AlexNet-based networks and demonstrates better level of performance. 
We achieved the classification accuracy of 0.933, 0.867 and 0.733 for binary AD/NC, AD/MCI and MCI/NC classification problems respectively and 0.689 for ternary AD/MCI/NC classification problem on a subset of ADNI database consisting of 531 subjects with sMRI and DTI scans. The obtained results make us think that the CNN-based classification with fusion of modalities can indeed be used for real-world CAD systems in large cohort screening.

In this paper we focused on only one biomarker ROI, the hippocampal ROI. Nevertheless, accordingly to previous research \cite{f3_Jenny_pcc} it is interesting to add other ROIs known to be deteriorated due to AD.

\section{Acknowledgements}
Data  collection  and  sharing  for  this  project  was  funded  by  the  Alzheimer's  Disease Neuroimaging  Initiative  (ADNI)  (National  Institutes  of  Health  Grant  U01  AG024904)  and DOD  ADNI  (Department  of  Defense  award  number  W81XWH-12-2-0012). ADNI  is  funded by  the  National Institute  on  Aging,  the  National  Institute of  Biomedical  Imaging  and Bioengineering, and through generous contributions from the following: AbbVie, Alzheimer’s Association;  Alzheimer’s  Drug  Discovery  Foundation; Araclon Biotech; BioClinica, Inc.; Biogen; Bristol-Myers Squibb Company; CereSpir, Inc.; Cogstate;Eisai Inc.; Elan Pharmaceuticals, Inc.; Eli Lilly and Company; EuroImmun;  F.Hoffmann-La  Roche  Ltd  and its  affiliated  company  Genentech, Inc.;  Fujirebio;  GE Healthcare;  IXICO  Ltd.;  Janssen Alzheimer    Immunotherapy  Research  and   Development, LLC.; Johnson   and  Johnson Pharmaceutical  Research  and Development LLC.; Lumosity; Lundbeck; Merck and Co., Inc.; Meso Scale Diagnostics, LLC.; NeuroRx  Research; Neurotrack Technologies; Novartis Pharmaceuticals Corporation; Pfizer Inc.; Piramal Imaging; Servier; Takeda Pharmaceutical Company; and Transition Therapeutics.
The  Canadian  Institutes  of  Health  Research  is providing  funds  to  support  ADNI  clinical  sites  in  Canada.  Private  sector  contributions  are facilitated by the Foundation for the National Institutes of Health (\url{www.fnih.org}). The grantee organization is the Northern California Institute for Research and Education, and the study is coordinated by the Alzheimer’s Therapeutic Research Institute at the University of Southern California.  ADNI  data  are  disseminated  by  the  Laboratory  for  Neuroimaging  at  the University of Southern California.

This research was supported by Ostrogradsky scholarship grant 2017 established by French Embassy in Russia and TOUBKAL French-Morocco research grant Alclass. We thank Dr. Pierrick Coup\'e from LABRI UMR 5800 University of Bordeaux/CNRS/Bordeaux-IPN who provided insight and expertise that greatly assisted the research.

\bibliographystyle{IEEEtranN}
\bibliography{biblio}

\end{document}